\documentclass[11pt]{article}

\usepackage[preprint]{acl}

\usepackage{times}
\usepackage{latexsym}

\usepackage[T1]{fontenc}

\usepackage[utf8]{inputenc}

\usepackage{microtype}

\usepackage{inconsolata}

\usepackage{graphicx}
\usepackage[T1]{fontenc}    
\usepackage{hyperref}       
\usepackage{url}            
\usepackage{booktabs}       
\usepackage{amsfonts}       
\usepackage{nicefrac}       
\usepackage{xcolor}         
\usepackage{geometry}
\usepackage{amsmath}
\usepackage{multirow}
\usepackage{enumitem}
\title{\textsc{DFlare}: Scaling Up Draft Capacity for Block Diffusion Speculative Decoding}

\author{%
  \bfseries 
  Jiebin Zhang$^{1}$ \quad 
  Zhenghan Yu$^{1}$ \quad 
  Song Liu$^{2}$ \quad 
  Eugene J. Yu$^{1}$ \quad
  Zheng Li$^{1}$ \quad 
  Dawei Zhu$^{1}$ \\[0.5ex]
  \bfseries 
  Jiangshan Duo$^{1}$ \quad 
  Weimin Xiong$^{1}$ \quad 
  Yifan Song$^{1}$ \quad 
  Guanghua Yu$^{2}$ \quad 
  Jianchen Zhu$^{2}$ \quad 
  Sujian Li$^{1}$ \\[1.5ex]
  \normalfont $^1$School of Computer Science, Peking University \quad $^2$Tencent \\ 
  \small \texttt{\{zhangjiebin, lisujian\}@pku.edu.cn} \quad \texttt{lucayu@tencent.com}
}

\begin{document}
\newcommand{\modelname}{\textsc{DFlare} }
\maketitle
\begin{abstract}
Block diffusion speculative decoding accelerates LLM inference by predicting all tokens within a block simultaneously for the target model to verify in parallel. Predicting an entire block at once requires a sufficiently capable draft model and effective utilization of the target model's internal knowledge. However, the state-of-the-art method DFlash constrains all draft layers to share a single fused representation derived from only a few target layers, limiting per-layer expressiveness and hindering further scaling of draft capacity. In this paper, we present \modelname, which flares out the narrow conditioning bottleneck of DFlash through a lightweight layer-wise fusion mechanism: 
each draft layer attends to its own learnable combination of a broad set of target layers at negligible overhead, simultaneously injecting richer target knowledge and providing every draft layer with a distinct input. This enhanced per-layer expressiveness enables scaling the draft model to deeper architectures with consistent gains. We further scale training data from 800K to 2.4M samples to fully exploit the enlarged capacity. On six benchmarks spanning mathematical reasoning, code generation, and conversation, \modelname  attains average wall-clock speedups of $5.52\times$ on Qwen3-4B, $5.46\times$ on Qwen3-8B, and $3.91\times$ on GPT-OSS-20B, improving over DFlash by roughly $11\%$, $8\%$, and $5\%$ respectively. Our code is
available at \href{https://github.com/Tencent/AngelSlim}{https://github.com/Tencent/AngelSlim}.
\end{abstract}

\begin{figure*}[t]
  \centering
  \includegraphics[width=\textwidth]{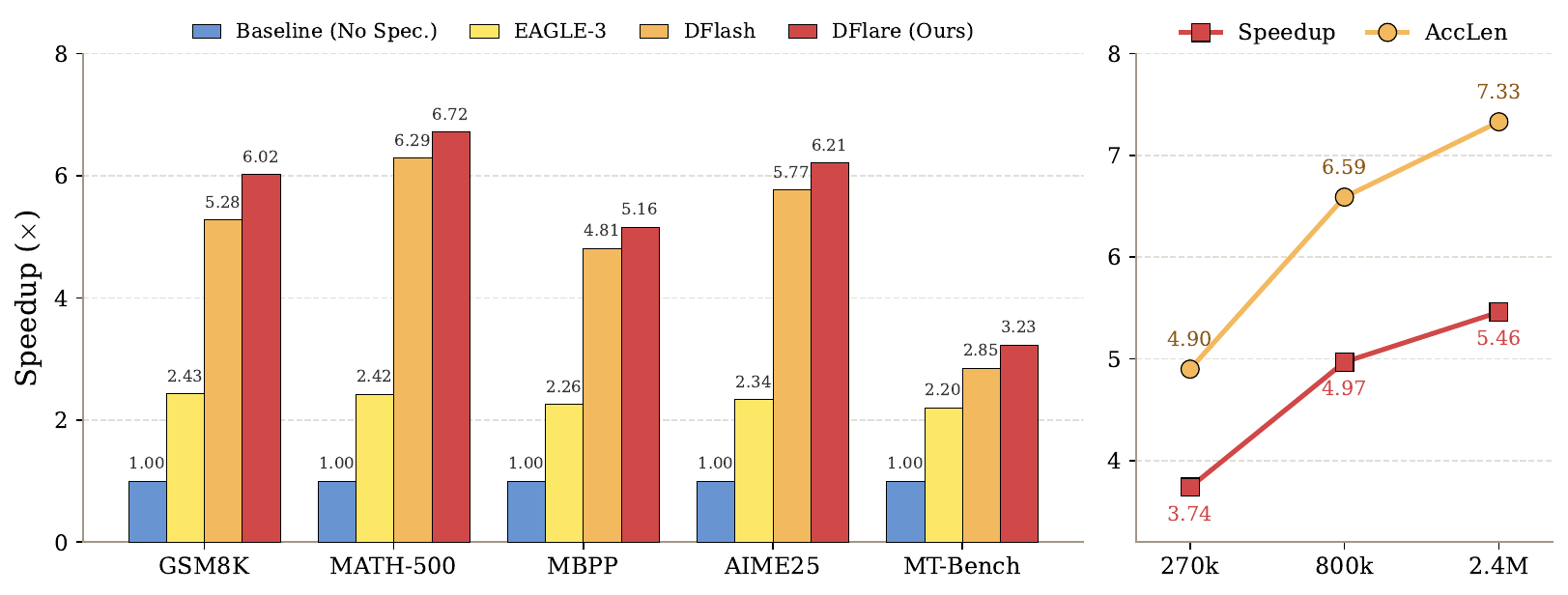}
   \caption{Left: wall-clock speedup ($\times$) of different speculative decoding methods on Qwen3-8B across five benchmarks under greedy decoding; \modelname consistently achieves the highest speedup, outperforming DFlash by a significant margin on every benchmark. Right: speedup and acceptance length of \modelname on Qwen3-8B as the training data scales from 270k to 2.4M samples; \modelname delivers consistent and substantial improvements as more training data becomes available.}
  \label{fig:speedup}
\end{figure*}

\section{Introduction}

Speculative decoding~\citep{leviathan2023fast, chen2023accelerating} accelerates LLM inference by employing a lightweight \emph{draft model} to predict multiple future tokens, which the larger \emph{target model} then verifies in a single parallel forward pass. This preserves the target distribution while substantially reducing wall-clock time~\citep{xia-etal-2024-unlocking}. The resulting speedup critically depends on the \emph{acceptance length}---the number of draft tokens accepted per verification step---which reflects how well the draft model approximates the target model~\citep{liu2023online,zhou2023distillspec}. Early approaches predominantly adopt autoregressive draft models~\citep{leviathan2023fast, chen2023accelerating, li2024eagle, li2025eagle}, but their sequential token-by-token generation means that each additional layer directly increases per-token latency, severely constraining the draft model's size and capacity.

Recently, \textit{block diffusion} drafting~\citep{sandler2025specdiff2scalingdiffusiondrafter,li2025diffuspecunlockingdiffusionlanguage,chen2026dflashblockdiffusionflash} has introduced a new paradigm: instead of generating candidate tokens autoregressively, it predicts all tokens within a block simultaneously through a discrete diffusion process in a single forward pass whose latency grows only marginally with the number of predicted tokens. 
This fundamentally changes the design space for draft models: since drafting cost is largely decoupled from block size,  draft architectures with greater capability  become practically viable and desirable for predicting more tokens within a block simultaneously.

To improve draft quality, two complementary strategies are commonly employed: (1)~\textit{scaling draft depth} to enhance  modeling capacity~\citep{du2024glide,yan2025scalinglawsspeculativedecoding}, which requires that the acceptance length gain from each additional layer outweighs the extra computational cost it introduces; and (2)~\textit{injecting target knowledge} by conditioning the draft model on the internal representations of the target model ~\citep{cai2024medusa,li2024eagle,li2025eagle}, which requires the draft model to effectively leverage the injected information to boost acceptance length.
To the best of our knowledge, research along these directions for block diffusion drafting remains preliminary.
For instance, the current state-of-the-art method DFlash \citep{chen2026dflashblockdiffusionflash} has not fully exploited the potential of either improvement axis above.
On the depth axis, DFlash employs only a modest number of draft layers and exhibits diminishing returns when more layers are added. On the knowledge injection axis, DFlash fuses hidden states from a few target layers into a single representation and shares it across all draft layers. Feeding the \emph{same} fused signal to every layer prevents individual layers from specializing, which also explains why depth scaling saturates:
without distinct input, additional layers have little room to contribute new expressiveness. Importantly, the two limitations are tightly coupled, unlocking one requires addressing the other.

To jointly address these coupled limitations, 
we present \modelname, which flares out the narrow conditioning bottleneck of DFlash through broadening the target-model information flow from a single shared representation into distinct, layer-wise conditioning signals. 
Specifically, we introduce a \textit{lightweight layer-wise fusion mechanism}: each draft layer learns its own weighted combination over target hidden states via a simple scalar-weighted sum, providing every layer with a distinct conditioning signal that encourages specialization. This differentiated input breaks the saturation ceiling, allowing each additional draft layer to contribute meaningful gains rather than redundant computation. Simultaneously, this mechanism enriches target knowledge injection by drawing from substantially more target layers than DFlash (e.g., 9 vs.\ 5) with minimal overhead, since the fusion requires only a handful of scalar weights per layer, adding negligible computation. The two improvements reinforce each other: richer per-layer knowledge makes depth scaling effective, while greater depth in turn amplifies the utility of the injected information. To fully exploit the enlarged model capacity, we further scale training data from 800K to 2.4\,M samples, providing the supervision needed for the deeper, more expressive model to converge. On six benchmarks spanning mathematical reasoning, code generation, and general conversation, \modelname attains average wall-clock speedups of $5.52\times$ on Qwen3-4B, $5.46\times$ on Qwen3-8B, and $3.91\times$ on GPT-OSS-20B, improving over DFlash by roughly $11\%$, $8\%$, and $5\%$ respectively, as shown in Figure~\ref{fig:speedup} Left.


Our main contributions are as follows:
\begin{itemize}
    \item \textbf{Scaling target knowledge per layer.} We introduce a lightweight layer-wise fusion mechanism that lets each draft layer attend to its own learnable combination of target layers at negligible cost. This allows incorporating far more target layers than prior work (e.g., 9 vs.\ 5 in DFlash), providing richer conditioning while giving every draft layer a differentiated view to strengthen per-layer expressiveness.
    \item \textbf{Scaling draft depth.} Building on the enhanced per-layer expressiveness, we scale the draft model to more layers and demonstrate that acceptance length improves consistently with depth---confirming that layer-wise fusion effectively unlocks the depth-scaling potential of block diffusion.
    \item \textbf{Scaling training data.} We scale the training corpus from 800K to 2.4\,M samples to match the enlarged capacity, the results are shown in Figure~\ref{fig:speedup} Right. \modelname keeps improving as more data is supplied, and ultimately attains average wall-clock speedups of $5.52\times$ on Qwen3-4B, $5.46\times$ on Qwen3-8B, and $3.91\times$ on GPT-OSS-20B, improving over DFlash by roughly $11\%$, $8\%$, and $5\%$ respectively.

\end{itemize}

\section{Related Work}
\label{sec:related_work}

\subsection{Speculative Decoding}
Speculative decoding~\citep{leviathan2023fast, chen2023accelerating} accelerates large language model inference by introducing a \emph{draft-then-verify} paradigm~\citep{xia-etal-2024-unlocking}. Various methods aim to improve draft model quality within this autoregressive framework—for instance, by distilling the target model~\citep{zhou2023distillspec,liu2023online}, utilizing target model features~\citep{li2024eagle,zhang2025learningharmonizedrepresentationsspeculative,du2024glide,li2025eagle}, or employing tree-based drafting strategies~\citep{miao2023specinfer,li2024eagle2fasterinferencelanguage,hu2025bridging}.  Another line of work leverages parallel speculative decoding; for example, some approaches employ multiple token prediction heads for parallel prediction~\citep{cai2024medusa,10.5555/3692070.3692699}, while others directly leverage pre-trained diffusion models as draft model~\citep{christopher2025speculativediffusiondecodingaccelerating,sandler2025specdiff2scalingdiffusiondrafter,li2025diffuspecunlockingdiffusionlanguage,liu2025tidarthinkdiffusiontalk}. Notably, DFlash~\citep{chen2026dflashblockdiffusionflash} trains a block diffusion draft model conditioned on the target model's hidden states via KV injection, achieving state-of-the-art acceleration.

\subsection{Diffusion Language Models}
Diffusion language models generate text by iteratively denoising corrupted sequences~\citep{li2025survey}. Early works explored both continuous-space diffusion over token embeddings~\citep{li2022diffusionlm,strudel2022selfconditioned} and discrete-space diffusion with structured transition matrices~\citep{austin2023structured,he2022diffusionbert}. Recent advances in discrete diffusion have substantially improved training objectives and scalability through simplified masked diffusion losses and score-based formulations~\citep{sahoo2024simple,shi2025simplified}. Block diffusion~\citep{arriola2025block,wu2025fastdllmtrainingfreeaccelerationdiffusion,wu2025fastdllmv2efficientblockdiffusion} combines this parallelism with autoregressive structure by denoising sequences block-by-block, where tokens within each block are generated in parallel while blocks are produced sequentially. 




\section{Preliminaries}
To facilitate understanding of our \modelname method, this section presents the key components of DFlash, upon which our approach builds. We first describe how target-model information is extracted and injected into the draft model (Section~\ref{sec:prelim_target_info}), followed by the inference procedure encompassing the draft-and-verification loop (Section~\ref{sec:prelim_inference}). Together, these two aspects illuminate the architecture of the block diffusion draft model and motivate the specific improvements introduced by our method. Finally, we describe the training procedure of the block diffusion draft model (Section~\ref{sec:prelim_training}), which our method also adopts.

\subsection{Leveraging target model information}
\label{sec:prelim_target_info}

The draft model, with its limited capacity, struggles to precisely approximate the output distribution of the large-scale target model when drafting from scratch. A key insight in recent speculative decoding methods is to leverage the target model’s internal hidden states as additional context for the draft model~\citep{li2024eagle,li2025eagle}. DFlash~\citep{chen2026dflashblockdiffusionflash} implements this idea by fusing hidden states from multiple layers of the target model via a fully-connected (FC) layer.

Concretely, let $\mathbf{H}^{(j)} \in \mathbb{R}^{L \times d}$ ($j = 1, \ldots, T$) denote the hidden state matrix from the $j$-th uniformly sampled target layer across all $L$ context positions, where $d$ is the hidden dimension. At each position $t$, the hidden states from all $T$ layers are concatenated and projected through an FC layer to produce a context feature:
\begin{equation}
  \mathbf{c}_t = \mathbf{W}_{\text{fc}} [\mathbf{h}_t^{(1)}; \ldots; \mathbf{h}_t^{(T)}] + \mathbf{b}_{\text{fc}}, \quad \mathbf{c}_t \in \mathbb{R}^{d},
  \label{eq:dflash_fc}
\end{equation}
where $\mathbf{h}_t^{(j)} \in \mathbb{R}^{d}$ is the $t$-th row of $\mathbf{H}^{(j)}$, $\mathbf{W}_{\text{fc}} \in \mathbb{R}^{d \times Td}$, and $\mathbf{b}_{\text{fc}} \in \mathbb{R}^{d}$. The context feature $\mathbf{c}_t$ integrates semantic information from different depths of the target model’s hierarchy. DFlash directly injects these context features into the Key and Value projections of \emph{every} draft model layer through the KV cache, providing persistent target model conditioning throughout the draft model’s depth (detailed in Section~\ref{sec:prelim_inference}). A notable limitation of this design is that all draft layers receive the \emph{identical} context feature $\mathbf{c}_t$, preventing different draft layers from specializing on different aspects of the target model’s representation.

\subsection{Inference of block diffusion draft model}
\label{sec:prelim_inference}

At inference time, the block diffusion draft model and the target model operate in an iterative draft-then-verify loop. Each iteration proceeds in two phases.

\paragraph{Drafting phase.} Let the target decode token at position $t$ denote the latest token produced by the target model. The draft model takes this token as the starting point and generates $B-1$ candidate tokens in parallel for positions $t{+}1, \ldots, t{+}B{-}1$. To condition the draft model on target model knowledge, the fused context features $\mathbf{c}_t$ (Eq.~\ref{eq:dflash_fc}) computed from all preceding positions---which we refer to as the \emph{target context tokens}---are injected into the KV cache of every draft layer. Let $\mathbf{C}$ denote the matrix stacking the context features $\mathbf{c}_t$ for all $L$ context positions. In each draft layer, the Keys and Values are formed by concatenating the projections from three sources: (1) the target context features $\mathbf{C}$ for $L$ context positions, (2) the target decode token representation $\mathbf{x}_t \in \mathbb{R}^{1 \times d}$, and (3) the draft masked-position hidden states $\hat{\mathbf{X}} \in \mathbb{R}^{(B-1) \times d}$:
\begin{equation}
\begin{split}
  \mathbf{K} &= \bigl[\mathbf{C}\, \mathbf{W}_K \;;\; \mathbf{x}_t\, \mathbf{W}_K \;;\; \hat{\mathbf{X}}\, \mathbf{W}_K\bigr], \\
  \mathbf{V} &= \bigl[\mathbf{C}\, \mathbf{W}_V \;;\; \mathbf{x}_t\, \mathbf{W}_V \;;\; \hat{\mathbf{X}}\, \mathbf{W}_V\bigr],
\end{split}
\label{eq:baseline_attn}
\end{equation}
with Queries $\mathbf{Q} = [\mathbf{x}_t;\, \hat{\mathbf{X}}]\, \mathbf{W}_Q$, where $\mathbf{W}_K, \mathbf{W}_V, \mathbf{W}_Q$ are the standard projection matrices of the draft model. Note that all three sources share the \emph{same} KV projections. The attention is \emph{bidirectional} within the block: every position attends to all other block positions and to the full target context, allowing the model to resolve inter-token dependencies in a single forward pass. The attention output is then processed by the MLP sub-layer (following the standard Transformer block structure). Only the $B$ block positions (i.e., $\mathbf{x}_t$ and $\hat{\mathbf{X}}$) are passed to subsequent draft layers; the target context features $\mathbf{C}$ remain fixed across layers. After the final draft layer, the language modeling head produces logits at each masked position, from which $B-1$ candidate tokens are sampled.

\paragraph{Verification phase.} The target decode token together with the $B-1$ candidates are fed into the target model for parallel verification. The target model evaluates all $B$ positions in a single forward pass and accepts the longest prefix whose tokens match its own predictions. In addition to the accepted tokens, the target model generates one extra token at the position immediately following the last accepted token. This newly generated token serves as the target decode token for the next drafting iteration, and the target model's hidden states and KV cache are updated accordingly. This cycle repeats until generation is complete.

\subsection{Training for block diffusion draft model}
\label{sec:prelim_training}


During training, multiple \emph{anchor} positions are randomly sampled from the response. Each anchor serves as the target decode token $\mathbf{x}_t$ of a block, and the subsequent $B-1$ positions are masked as $\hat{\mathbf{X}}$---directly aligned with the inference procedure described in Section~\ref{sec:prelim_inference}. All sampled blocks are concatenated into a single sequence and processed jointly using a sparse attention mask: tokens attend bidirectionally within the same block and to the corresponding target context features, while attention across different blocks is disallowed. This design enables multiple draft blocks to be trained efficiently within a single forward and backward pass. To improve training efficiency, the draft model shares the token embedding layer and the language modeling head with the target model and keeps them frozen; only the draft Transformer layers are updated.

\section{Method}
\label{sec:method}

\begin{figure*}[t]
    \centering
    \includegraphics[width=\linewidth]{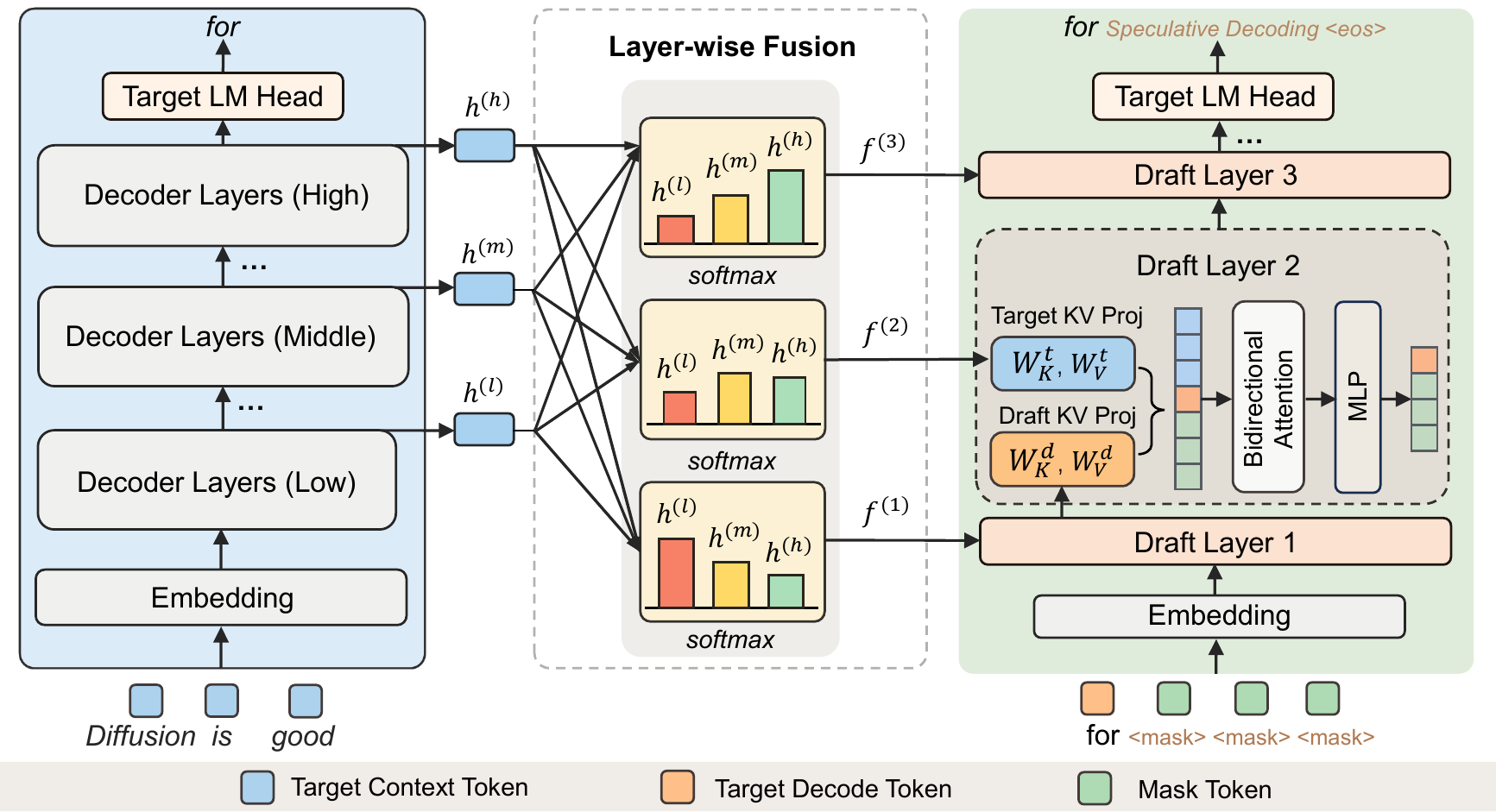}
    \caption{The overview of our \modelname method. \modelname utilizes adaptive layer fusion of target hidden states and heterogeneous KV projections to enhance per-layer expressiveness.}
    \label{fig:intro}
\end{figure*}

Building upon DFlash (Section~\ref{sec:prelim_target_info}--\ref{sec:prelim_training}), we present \modelname with three improvements that collectively \emph{strengthen the per-layer capacity} of the draft model: (1) a lightweight layer-wise fusion mechanism that replaces the FC projection, allowing each draft layer to receive its own dedicated combination of target hidden states and thereby maximizing the expressiveness of every single layer; (2) heterogeneous KV projections that decouple the representational spaces for draft and target information, granting each layer additional degrees of freedom to independently extract and utilize target knowledge; and (3) a progressive position-weighted training loss that improves training efficiency so the enlarged per-layer capacity is fully exploited. Together, these designs ensure that each draft layer becomes a more powerful computational unit, enabling the model to scale to greater depth with consistent gains. We describe each component below.

\subsection{Adaptive Layer Fusion of Target Hidden States}
\label{sec:layer_fusion}

Different layers of the target model encode distinct levels of abstraction—from shallow syntactic patterns to deep semantic representations. To effectively inject this multi-granularity knowledge into the draft model, \modelname introduces \emph{Adaptive Layer Fusion}: a lightweight, layer-specific mechanism that allows each draft layer to learn its own combination of target hidden states. Unlike prior approaches that project the concatenated target states through an FC layer to produce a single, shared fused context for all draft layers (Eq.~\ref{eq:dflash_fc}), our method provides each draft layer with a dedicated view of the target model.

Concretely, let $\mathbf{h}_t^{(j)} \in \mathbb{R}^d$ denote the hidden state at position $t$ from the $j$-th selected target layer ($j = 1, \ldots, T$). We introduce a learnable fusion weight matrix $\mathbf{W}^{\text{fuse}} \in \mathbb{R}^{D \times T}$, where $D$ is the number of draft layers. For each draft layer $i$, a layer-specific fused representation is computed via:
\begin{equation}
    \begin{split}
        \boldsymbol{\alpha}^{(i)} = \mathrm{softmax}\!\bigl(\mathbf{W}^{\text{fuse}}_{i,:}\bigr) \in \mathbb{R}^T, \\
  \mathbf{f}_t^{(i)} = \mathrm{RMSNorm}\!\!\left(\sum_{j=1}^{T} \alpha_j^{(i)} \, \mathbf{h}_t^{(j)}\right),
    \end{split}
  \label{eq:layer_fusion}
\end{equation}
where $\alpha_j^{(i)}$ is the softmax-normalized weight that draft layer $i$ assigns to target layer $j$. Once training is complete, the softmax-normalized weights $\boldsymbol{\alpha}^{(i)}$ can be precomputed and cached. At inference time, the fusion reduces to a scalar-weighted sum of the $T$ target hidden states per draft layer, introducing virtually no additional latency.

This design offers two key advantages. First, it is \emph{extremely lightweight}: the fusion introduces only $D \times T$ scalar parameters (e.g., $63$ for $D{=}7, T{=}9$), incurring only minimal additional overhead when more target layers are incorporated—allowing the draft model to absorb richer knowledge from the target model. Second, it is \emph{layer-specific}: each draft layer learns its own fusion coefficients, enabling each draft layer to receive information suited to its role in the model hierarchy.

\subsection{Heterogeneous KV Projections}
\label{sec:hetero_kv}

The fused target context $\mathbf{f}_t^{(i)}$ is injected into each draft layer through the Key-Value (KV) cache, supplying the draft model with information from the target model. In DFlash, the draft model's own hidden states and the injected target context share the same KV projection matrices (Eq.~\ref{eq:baseline_attn}). However, these two sources carry different information: the draft hidden states capture the evolving noisy predictions within the diffusion process, while the target context encodes semantic knowledge from the target model. Sharing a single projection forces the model to compromise between two distinct representational spaces. \modelname addresses this by introducing separate projection matrices for each source. Let $\mathbf{F}^{(i)} \in \mathbb{R}^{L \times d}$ denote the layer-specific fused target context at draft layer $i$ (obtained via Eq.~\ref{eq:layer_fusion}), $\mathbf{x}_t \in \mathbb{R}^{1 \times d}$ the target decode token representation, and $\hat{\mathbf{X}} \in \mathbb{R}^{(B-1) \times d}$ the draft masked-position hidden states. The Keys and Values are computed and concatenated as:
\begin{equation}
    \begin{split}
          \mathbf{K} = \bigl[\mathbf{F}^{(i)} \mathbf{W}_K^{t} \;;\; \mathbf{x}_t\, \mathbf{W}_K^{d} \;;\; \hat{\mathbf{X}}\, \mathbf{W}_K^{d}\bigr],  \\
  \mathbf{V} = \bigl[\mathbf{F}^{(i)} \mathbf{W}_V^{t} \;;\; \mathbf{x}_t\, \mathbf{W}_V^{d} \;;\; \hat{\mathbf{X}}\, \mathbf{W}_V^{d}\bigr],
    \end{split}
  \label{eq:hetero_kv}
\end{equation}
where $\mathbf{W}_K^{t}, \mathbf{W}_V^{t} \in \mathbb{R}^{d \times d_{\text{kv}}}$ are learnable projections dedicated to the target context, and $\mathbf{W}_K^{d}, \mathbf{W}_V^{d} \in \mathbb{R}^{d \times d_{\text{kv}}}$ are independent projections shared by the target decode token and the masked draft positions. The Queries are derived from the block positions: $\mathbf{Q} = [\mathbf{x}_t;\, \hat{\mathbf{X}}]\, \mathbf{W}_Q$. This heterogeneous design grants the target context and the draft block positions independent representational subspaces in the attention mechanism, enabling the draft model to better extract and utilize the target knowledge. The resulting Keys and Values are then used with the bidirectional attention mechanism described in Section~\ref{sec:prelim_inference} to compute the attention output.

\begin{table*}[t]                                                                                                                      

\centering                                                                                                                            
\resizebox{\textwidth}{!}{%

\setlength{\tabcolsep}{1pt}

\begin{tabular}{*{16}{c}}
\toprule
\multirow{2}{*}{\normalsize{Model}} & \multirow{2}{*}{\normalsize{Method}} & \multicolumn{6}{c}{MATH} & \multicolumn{4}{c}{CODE} & \multicolumn{2}{c}{CHAT} & \multicolumn{2}{c}{\multirow{2}{*}{Avg.}} \\

\cmidrule(lr){3-14}

& & \multicolumn{2}{c}{\textit{GSM8K}} & \multicolumn{2}{c}{\textit{MATH500}} & \multicolumn{2}{c}{\textit{AIME25}} & \multicolumn{2}{c}{\textit{HumanEval}} & \multicolumn{2}{c}{\textit{MBPP}} & \multicolumn{2}{c}{\textit{MT-Bench}} & \multicolumn{2}{c}{} \\
\midrule

\multicolumn{2}{c}{Temperature = 0} & \textit{Speedup} & \textit{$\tau$} & \textit{Speedup} & \textit{$\tau$} & \textit{Speedup} &
\textit{$\tau$} & \textit{Speedup} & \textit{$\tau$} & \textit{Speedup} & \textit{$\tau$} & \textit{Speedup} & \textit{$\tau$} &
\textit{Speedup} & \textit{$\tau$} \\

\midrule

\multirow{3}{*}{Q3-4B} & EAGLE3 & 2.56 & 3.76 & 2.57 & 3.70 & 2.46 & 3.58 & 2.52 & 3.58 & 2.41 & 3.51 & 2.39 & 3.57 & 2.49 & 3.62 \\

& DFlash & 5.31 & 6.51 & 6.20 & 7.80 & 5.78 & 7.31 & 4.62 & 6.65 & 4.89 & 6.13 & 3.13 & 4.40 & 4.99 & 6.47 \\                        

& \modelname & \textbf{6.00} & \textbf{7.93} & \textbf{6.76} & \textbf{8.97} & \textbf{6.31} & \textbf{8.35} & \textbf{5.41} &       
\textbf{7.17} & \textbf{5.30} & \textbf{7.08} & \textbf{3.33} & \textbf{5.30} & \textbf{5.52} & \textbf{7.47} \\

\midrule

\multirow{3}{*}{Q3-8B} & EAGLE3 & 2.43 & 3.53 & 2.42 & 3.55 & 2.34 & 3.51 & 2.46 & 3.67 & 2.26 & 3.31 & 2.20 & 3.25 & 2.35 & 3.47 \\

& DFlash & 5.28 & 6.53 & 6.29 & 7.88 & 5.77 & 7.11 & 5.32 & 6.54 & 4.81 & 5.97 & 2.85 & 4.30 & 5.05 & 6.39 \\

& \modelname & \textbf{6.02} & \textbf{7.88} & \textbf{6.72} & \textbf{8.95} & \textbf{6.21} & \textbf{8.12} & \textbf{5.42} &
\textbf{7.08} & \textbf{5.16} & \textbf{6.86} & \textbf{3.23} & \textbf{5.09} & \textbf{5.46} & \textbf{7.33} \\

\midrule

\multirow{2}{*}{GPT-20B} & DFlash & 3.75 & 4.54 & 3.96 & 4.86 & 3.85 & 4.93 & 3.49 & 4.25 & 3.44 & 4.20 & 3.74 & 5.20 & 3.71 & 4.66 \\

& \modelname & \textbf{4.06} & \textbf{4.98} & \textbf{4.20} & \textbf{5.19} & \textbf{3.88} & \textbf{4.98} & \textbf{3.71} &
\textbf{4.55} & \textbf{3.72} & \textbf{4.58} & \textbf{3.87} & \textbf{5.32} & \textbf{3.91} & \textbf{4.93} \\

\midrule

\multicolumn{2}{c}{Temperature = 1} & \textit{Speedup} & \textit{$\tau$} & \textit{Speedup} & \textit{$\tau$} & \textit{Speedup} &
\textit{$\tau$} & \textit{Speedup} & \textit{$\tau$} & \textit{Speedup} & \textit{$\tau$} & \textit{Speedup} & \textit{$\tau$} &
\textit{Speedup} & \textit{$\tau$} \\

\midrule

\multirow{3}{*}{Q3-4B} & EAGLE3 & 2.37 & 3.71 & 2.43 & 3.57 & 2.14 & 3.26 & 2.36 & 3.57 & 2.30 & 3.48 & 2.22 & 3.49 & 2.30 & 3.51 \\

& DFlash & 4.80 & 6.00 & 5.22 & 6.75 & \textbf{4.02} & 5.18 & 4.85 & 6.04 & 4.50 & 5.61 & 2.69 & 4.03 & 4.35 & 5.60 \\

& \modelname & \textbf{5.38} & \textbf{7.14} & \textbf{5.33} & \textbf{7.34} & 3.84 & \textbf{5.20} & \textbf{4.87} & \textbf{6.43} &
\textbf{4.75} & \textbf{6.39} & \textbf{3.00} & \textbf{4.77} & \textbf{4.53} & \textbf{6.21} \\

\midrule

\multirow{3}{*}{Q3-8B} & EAGLE3 & 2.23 & 3.44 & 2.20 & 3.40 & 2.11 & 3.11 & 2.31 & 3.54 & 2.12 & 3.25 & 1.99 & 3.06 & 2.16 & 3.30 \\

& DFlash & 4.83 & 5.96 & 4.99 & 6.39 & 3.80 & 4.82 & \textbf{4.52} & 5.54 & 4.26 & 5.29 & 2.57 & 3.79 & 4.16 & 5.30 \\

& \modelname & \textbf{5.34} & \textbf{7.07} & \textbf{5.25} & \textbf{7.18} & \textbf{3.84} & \textbf{5.24} & 4.41 &
\textbf{5.76} & \textbf{4.31} & \textbf{5.77} & \textbf{2.81} & \textbf{4.37} & \textbf{4.33} & \textbf{5.90} \\

\midrule

\multirow{2}{*}{GPT-20B} & DFlash & 2.95 & 3.64 & 2.99 & 3.76 & 2.04 & 3.10 & 2.63 & 3.20 & 2.58 & 3.13 & 1.76 & 2.76 & 2.49 & 3.26 \\

& \modelname & \textbf{3.07} & \textbf{3.83} & \textbf{3.08} & \textbf{3.91} & \textbf{2.05} & \textbf{3.14} & \textbf{2.67} &
\textbf{3.28} & \textbf{2.69} & \textbf{3.30} & \textbf{1.81} & \textbf{2.89} & \textbf{2.56} & \textbf{3.39} \\

\bottomrule

\end{tabular}%
}
\caption{Main results on Qwen3-4B, Qwen3-8B and GPT-OSS-20B across mathematical reasoning, code generation, and conversation benchmarks under
greedy and stochastic decoding. Speedup denotes wall-clock speedup ratio and $\tau$ denotes acceptance length. Best results in each
section are in \textbf{bold}.}

\label{tab:main_results}

\end{table*}

\subsection{Progressive Position-Weighted Loss}
\label{sec:progressive_loss}

In speculative decoding, errors at early positions within a draft block invalidate all subsequent tokens, making early-position accuracy disproportionately important. DFlash addresses this by applying an exponentially decaying weight $w_k = \exp(-(k-1)/\gamma)$ to the cross-entropy loss at each position $k$ within a block, where $\gamma$ controls the decay rate. However, DFlash uses a fixed $\gamma$ throughout training: a small $\gamma$ concentrates learning on early positions, enabling fast convergence but under-optimizing harder tail positions; a large $\gamma$ spreads the weight more uniformly but slows convergence on the easy early positions. We resolve this tension with a simple linear warmup:
\begin{equation}
    \begin{split}
          \gamma(s) = \gamma_0 + \frac{s}{S}\,(\gamma_{\max} - \gamma_0), \\
  w_k(s) = \exp\!\left(-\frac{k-1}{\gamma(s)}\right),
    \end{split}
  \label{eq:progressive_weight}
\end{equation}
where $s$ is the current training step and $S$ is the total number of steps. In the early phase ($\gamma \approx \gamma_0$, small), the model focuses on mastering the easy first few positions. As $\gamma$ increases toward $\gamma_{\max}$, the weight distribution flattens and the model progressively shifts its effort to later, harder positions. This curriculum-style schedule ensures that by the end of training, the model has been thoroughly optimized across all block positions, yielding longer accepted sequences without sacrificing early-position accuracy.

\section{Experiments}
\subsection{Setup}
\label{subsec:setup}
\paragraph{Models and Evaluations} We conduct experiments on Qwen3-4B and Qwen3-8B~\citep{qwen3technicalreport} and GPT-OSS-20B~\citep{agarwal2025gpt} pretrained models. We evaluate tasks in three categories: Math: GSM8K~\citep{cobbe2021gsm8k}, MATH500~\citep{hendrycks2021MATH}, and AIME25~\citep{aime25}; Code: HumanEval~\citep{chen2021humaneval}, MBPP~\citep{austin2021mbpp}; Chat: MTBench~\citep{zheng2023mtbench}. More details are presented in Section~\ref{apd:baseline}.

\paragraph{Datasets} 
To provide a diverse set of training data, we
collect a mixture of around 2.4M samples from NVIDIA
Nemotron Post-Training Dataset V2~\citep{nvidia2025nvidianemotronnano2}, CodeAlpaca~\citep{codealpaca} and Step-3.5-Flash-SFT~\citep{huang2026step}. More details are presented in Section~\ref{apd:ablation}.

\paragraph{Implementation}
For \modelname draft models, we set the
number of draft layers to 7 and use a block
size of 16. The target hidden features
are extracted from 9 layers uniformly selected between the
second layer and the third-to-last layer of the target model (8, 8, 7 for GPT-OSS).
More details are presented in Section~\ref{apd:train} and Section~\ref{apd:baseline}.
\subsection{Main Results}

Table~\ref{tab:main_results} reports the main results across six benchmarks spanning mathematical reasoning, code generation, and general conversation, evaluated on three target models: Qwen3-4B, Qwen3-8B, and GPT-OSS-20B. \modelname consistently outperforms DFlash across all benchmarks, target models, and temperature settings. Under greedy decoding, \modelname improves the average acceptance length by $15.5\%$ on Qwen3-4B, $14.7\%$ on Qwen3-8B, and $5.8\%$ on GPT-OSS-20B, with corresponding wall-clock speedup gains of $10.6\%$, $8.1\%$, and $5.4\%$ respectively.  The improvements vary across domains. Mathematical reasoning benchmarks benefit the most, code generation benchmarks also show substantial improvements. Conversational tasks exhibit relatively lower absolute acceptance lengths due to the diversity of open-ended generation, yet \modelname still delivers the largest \emph{relative} gain on MT-Bench. For the substantially larger GPT-OSS-20B target, both DFlash and \modelname attain lower absolute acceptance lengths than on the Qwen3 targets, which is expected given the larger representational gap the draft model must bridge; nevertheless, \modelname retains a consistent advantage over DFlash across every benchmark. Under stochastic decoding, acceptance lengths decrease across all methods as sampling introduces additional mismatch between draft and target distributions, but \modelname maintains clear margins of $10.9\%$, $11.3\%$, and $3.9\%$ in average acceptance length on Qwen3-4B, Qwen3-8B, and GPT-OSS-20B respectively, demonstrating the robustness of the proposed method.

\section{Analysis}

\subsection{Number of Draft Layers}
\label{sec:draft_layers}

\begin{table}[t] 
\centering
\small
\setlength{\tabcolsep}{3pt} 
\resizebox{\columnwidth}{!}{%
\begin{tabular}{l cc cc cc cc}
\toprule
\multirow{2}{*}{Setting} & \multicolumn{2}{c}{GSM8k} & \multicolumn{2}{c}{Humaneval} & \multicolumn{2}{c}{MtBench} & \multicolumn{2}{c}{Avg} \\
\cmidrule(lr){2-3} \cmidrule(lr){4-5} \cmidrule(lr){6-7} \cmidrule(lr){8-9}
 & Speedup & $\tau$ & Speedup & $\tau$ & Speedup & $\tau$ & Speedup & $\tau$ \\
\midrule
\modelname & \textbf{3.80} & \textbf{4.94} & 3.55 & 4.62 & 2.76 & 4.00 & \textbf{3.37} & \textbf{4.52} \\
$-$ Softmax & 3.70 & 4.85 & 3.45 & 4.53 & 2.77 & 4.02 & 3.31 & 4.47 \\
$-$ KVProj & 3.77 & 4.90 & 3.50 & 4.54 & 2.78 & 3.99 & 3.35 & 4.48 \\
$-$ Loss & 3.72 & 4.84 & \textbf{3.56} & \textbf{4.63} & 2.73 & 3.93 & 3.34 & 4.47 \\
$+$ FC & 3.74 & 4.87 & 3.53 & 4.59 & \textbf{2.78} & \textbf{4.06} & 3.35 & 4.51 \\
\bottomrule
\end{tabular}
}
\caption{Ablation on model structure design with Qwen3-4B as target model. Best results in each column are in \textbf{bold}.}
\label{tab:structure_ablation}
\end{table}

\begin{figure*}[t]
  \centering
  \includegraphics[width=\textwidth]{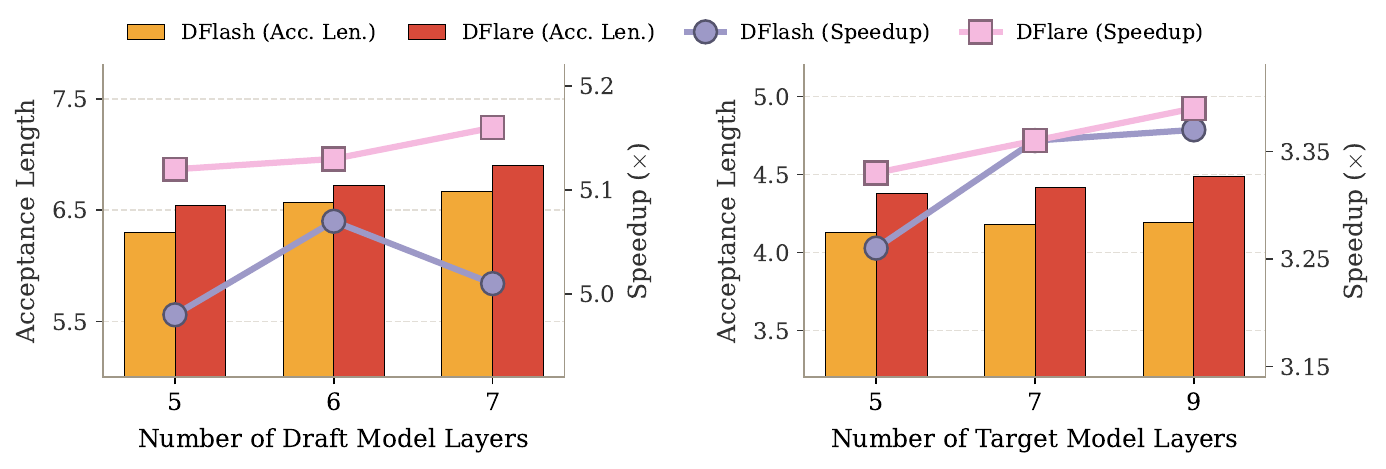}
  \caption{The impact of draft model layers (Left) and target model layers (Right) on the performance of DFlash and \modelname. The target model is Qwen3-4B.}
  \label{fig:layer}
\end{figure*}

We investigate how the number of draft model layers affects acceptance length and end-to-end speedup under our proposed architecture. To enable a fair per-layer comparison between \modelname and DFlash, we train both methods on the full DFlash training dataset and evaluate draft models with 5, 6, and 7 layers under otherwise identical settings. The results are visualized in Figure~\ref{fig:layer} Left.

\modelname consistently outperforms DFlash in both acceptance length and wall-clock speedup at every layer configuration, confirming that our layer-wise fusion mechanism strengthens the expressiveness of each individual draft layer. More importantly, the two methods exhibit qualitatively different scaling behaviors as draft depth increases. For \modelname, both acceptance length and speedup improve steadily from 5 to 7 layers---each additional layer yields a meaningful gain, indicating that the model continues to benefit from increased depth. In \modelname, each draft layer attends to its own dedicated combination of target hidden states, ensuring that every additional layer introduces genuinely new representational capacity. The consistent scaling of acceptance length with depth confirms that our layer-wise fusion effectively unlocks the depth scaling potential inherent to block diffusion drafting.

\subsection{Number of Target Hidden Features}
\label{sec:analysis_features}

We investigate how the number of fused target layers $T$ affects performance, and compare behavior of \modelname and DFlash as more target information is incorporated. Both methods are evaluated with $T \in \{5, 7, 9\}$ target layers under otherwise identical settings. The results are visualized in Figure~\ref{fig:layer} Right. \modelname demonstrates a clear ability to effectively leverage additional target layers. As $T$ increases from 5 to 9, acceptance length improves steadily and consistently, indicating that our lightweight layer-wise fusion successfully extracts complementary information from deeper target layers and translates it into higher-quality draft predictions. The speedup also improves monotonically, confirming that the richer target knowledge leads to longer accepted sequences that more than compensate for any marginal increase in per-step cost. In contrast, while increasing from 5 to 7 target layers provides a modest improvement, further scaling to 9 layers yields almost no additional acceptance length gain on DFlash.  The divergence between the two methods highlights the advantage of \modelname's layer-wise fusion design.

\subsection{Model Structures Design}
\label{sec:analysis_structure}

We ablate the key architectural choices of \modelname to validate our structure design. Specifically, we consider four variants:  (1) \textbf{$-$Softmax}: removing the softmax normalization on the per-layer fusion weights in Section~\ref{sec:layer_fusion}, leaving the learned combination unnormalized; (2) \textbf{$-$KVProj}: removing the heterogeneous KV projections described in Section~\ref{sec:hetero_kv}, so that the draft model's own hidden states and the fused target context share the same KV projection matrices; (3) \textbf{$-$Loss}: removing the progressive position-weighted loss strategy described in Section~\ref{sec:progressive_loss}; and (4) \textbf{$+$FC}: appending an additional fully-connected layer after the layer-wise fusion to further transform the fused representation before injection. Table~\ref{tab:structure_ablation} reports the results. Among the ablations, removing softmax normalization causes the most significant degradation, indicating that constraining the fusion weights is essential for stable.  Removing the heterogeneous KV projections also leads to a consistent drop in both acceptance length and speedup. Furthermore, the removal of the progressive position-weighted loss also results in a performance drop, demonstrating that our specialized loss design effectively guides the model to achieve a higher acceptance length. Interestingly, adding an extra FC layer after fusion does not improve acceptance length on average.  Overall, the \modelname configuration achieves the best overall acceptance length and speedup, confirming that all design choices are well-motivated. 

\section{Conclusion}
We presented \modelname, a method that scales up draft model capacity for block diffusion speculative decoding along three axes: target knowledge injection, draft model depth, and training data. By enabling each draft layer to learn its own fusion over target hidden states, \modelname unlocks effective depth scaling and benefits consistently from additional training data. On six benchmarks spanning mathematical reasoning, code generation, and conversation, \modelname  attains average wall-clock speedups of $5.52\times$ on Qwen3-4B, $5.46\times$ on Qwen3-8B, and $3.91\times$ on GPT-OSS-20B, improving over state-of-the-art DFlash by roughly $11\%$, $8\%$, and $5\%$ respectively.

\section*{Limitations}

Our work has two main limitations. First, the training cost of \modelname is high due to the large draft model and the scaled training corpus. However, since training is a one-time cost while the resulting draft model is deployed for all subsequent inference requests, the cumulative time savings during serving can quickly amortize the additional training investment. Second, increasing the training data is likely to yield further gains in acceptance length, yet we were unable to explore this direction due to computational resource constraints. We leave the investigation of even larger-scale training to future work.


\bibliography{custom}

@inproceedings{leviathan2023fast,
  title={Fast inference from transformers via speculative decoding},
  author={Leviathan, Yaniv and Kalman, Matan and Matias, Yossi},
  booktitle={International Conference on Machine Learning},
  pages={19274--19286},
  year={2023},
  organization={PMLR}
}

@article{chen2023accelerating,
  title={Accelerating large language model decoding with speculative sampling},
  author={Chen, Charlie and Borgeaud, Sebastian and Irving, Geoffrey and Lespiau, Jean-Baptiste and Sifre, Laurent and Jumper, John},
  journal={arXiv preprint arXiv:2302.01318},
  year={2023}
}

@inproceedings{10.5555/3692070.3692699,
author = {Gloeckle, Fabian and Idrissi, Badr Youbi and Rozi\`{e}re, Baptiste and Lopez-Paz, David and Synnaeve, Gabriel},
title = {Better \& faster large language models via multi-token prediction},
year = {2024},
publisher = {JMLR.org},
booktitle = {Proceedings of the 41st International Conference on Machine Learning},
articleno = {629},
numpages = {29},
location = {Vienna, Austria},
series = {ICML'24}
}

@misc{liu2025tidarthinkdiffusiontalk,
      title={TiDAR: Think in Diffusion, Talk in Autoregression}, 
      author={Jingyu Liu and Xin Dong and Zhifan Ye and Rishabh Mehta and Yonggan Fu and Vartika Singh and Jan Kautz and Ce Zhang and Pavlo Molchanov},
      year={2025},
      eprint={2511.08923},
      archivePrefix={arXiv},
      primaryClass={cs.CL},
      url={https://arxiv.org/abs/2511.08923}, 
}

@inproceedings{cai2024medusa,
  title={MEDUSA: Simple LLM inference acceleration framework with multiple decoding heads},
  author={Cai, Tianle and Li, Yuhong and Geng, Zhengyang and Peng, Hongwu and Lee, Jason D and Chen, Deming and Dao, Tri},
  booktitle={Proceedings of the 41st International Conference on Machine Learning},
  pages={5209--5235},
  year={2024}
}

@misc{chen2026dflashblockdiffusionflash,
      title={DFlash: Block Diffusion for Flash Speculative Decoding}, 
      author={Jian Chen and Yesheng Liang and Zhijian Liu},
      year={2026},
      eprint={2602.06036},
      archivePrefix={arXiv},
      primaryClass={cs.CL},
      url={https://arxiv.org/abs/2602.06036}, 
}

@misc{christopher2025speculativediffusiondecodingaccelerating,
      title={Speculative Diffusion Decoding: Accelerating Language Generation through Diffusion}, 
      author={Jacob K Christopher and Brian R Bartoldson and Tal Ben-Nun and Michael Cardei and Bhavya Kailkhura and Ferdinando Fioretto},
      year={2025},
      eprint={2408.05636},
      archivePrefix={arXiv},
      primaryClass={cs.CL},
      url={https://arxiv.org/abs/2408.05636}, 
}

@misc{sandler2025specdiff2scalingdiffusiondrafter,
      title={SpecDiff-2: Scaling Diffusion Drafter Alignment For Faster Speculative Decoding}, 
      author={Jameson Sandler and Jacob K. Christopher and Thomas Hartvigsen and Ferdinando Fioretto},
      year={2025},
      eprint={2511.00606},
      archivePrefix={arXiv},
      primaryClass={cs.CL},
      url={https://arxiv.org/abs/2511.00606}, 
}

@misc{li2025diffuspecunlockingdiffusionlanguage,
      title={DiffuSpec: Unlocking Diffusion Language Models for Speculative Decoding}, 
      author={Guanghao Li and Zhihui Fu and Min Fang and Qibin Zhao and Ming Tang and Chun Yuan and Jun Wang},
      year={2025},
      eprint={2510.02358},
      archivePrefix={arXiv},
      primaryClass={cs.CL},
      url={https://arxiv.org/abs/2510.02358}, 
}

@inproceedings{li2024eagle, 
	author = {Yuhui Li and Fangyun Wei and Chao Zhang and Hongyang Zhang}, 
	title = {{EAGLE}: Speculative Sampling Requires Rethinking Feature Uncertainty}, 
	booktitle = {International Conference on Machine Learning},
	year = {2024}
}

@article{li2025eagle,
  title={Eagle-3: Scaling up inference acceleration of large language models via training-time test},
  author={Li, Yuhui and Wei, Fangyun and Zhang, Chao and Zhang, Hongyang},
  journal={arXiv preprint arXiv:2503.01840},
  year={2025}
}

@misc{wu2025fastdllmtrainingfreeaccelerationdiffusion,
      title={Fast-dLLM: Training-free Acceleration of Diffusion LLM by Enabling KV Cache and Parallel Decoding}, 
      author={Chengyue Wu and Hao Zhang and Shuchen Xue and Zhijian Liu and Shizhe Diao and Ligeng Zhu and Ping Luo and Song Han and Enze Xie},
      year={2025},
      eprint={2505.22618},
      archivePrefix={arXiv},
      primaryClass={cs.CL},
      url={https://arxiv.org/abs/2505.22618}, 
}

@misc{wu2025fastdllmv2efficientblockdiffusion,
      title={Fast-dLLM v2: Efficient Block-Diffusion LLM}, 
      author={Chengyue Wu and Hao Zhang and Shuchen Xue and Shizhe Diao and Yonggan Fu and Zhijian Liu and Pavlo Molchanov and Ping Luo and Song Han and Enze Xie},
      year={2025},
      eprint={2509.26328},
      archivePrefix={arXiv},
      primaryClass={cs.CL},
      url={https://arxiv.org/abs/2509.26328}, 
}

@inproceedings{
arriola2025block,
title={Block Diffusion: Interpolating Between Autoregressive and Diffusion Language Models},
author={Marianne Arriola and Aaron Gokaslan and Justin T Chiu and Zhihan Yang and Zhixuan Qi and Jiaqi Han and Subham Sekhar Sahoo and Volodymyr Kuleshov},
booktitle={The Thirteenth International Conference on Learning Representations},
year={2025},
url={https://arxiv.org/abs/2503.09573}
}

@misc{cobbe2021gsm8k,
      title={Training Verifiers to Solve Math Word Problems}, 
      author={Karl Cobbe and Vineet Kosaraju and Mohammad Bavarian and Mark Chen and Heewoo Jun and Lukasz Kaiser and Matthias Plappert and Jerry Tworek and Jacob Hilton and Reiichiro Nakano and Christopher Hesse and John Schulman},
      year={2021},
      eprint={2110.14168},
      archivePrefix={arXiv},
      primaryClass={cs.LG},
      url={https://arxiv.org/abs/2110.14168}, 
}

@misc{hendrycks2021MATH,
      title={Measuring Mathematical Problem Solving With the MATH Dataset}, 
      author={Dan Hendrycks and Collin Burns and Saurav Kadavath and Akul Arora and Steven Basart and Eric Tang and Dawn Song and Jacob Steinhardt},
      year={2021},
      eprint={2103.03874},
      archivePrefix={arXiv},
      primaryClass={cs.LG},
      url={https://arxiv.org/abs/2103.03874}, 
}

@misc{aime25,
      title={American Invitational Mathematics Examination (AIME) 2025}, 
      author={Zhang, Yifan and Math-AI, Team},
      year={2025},
}

@misc{chen2021humaneval,
      title={Evaluating Large Language Models Trained on Code}, 
      author={Mark Chen and Jerry Tworek and Heewoo Jun and Qiming Yuan and Henrique Ponde de Oliveira Pinto and Jared Kaplan and Harri Edwards and Yuri Burda and Nicholas Joseph and Greg Brockman and Alex Ray and Raul Puri and Gretchen Krueger and Michael Petrov and Heidy Khlaaf and Girish Sastry and Pamela Mishkin and Brooke Chan and Scott Gray and Nick Ryder and Mikhail Pavlov and Alethea Power and Lukasz Kaiser and Mohammad Bavarian and Clemens Winter and Philippe Tillet and Felipe Petroski Such and Dave Cummings and Matthias Plappert and Fotios Chantzis and Elizabeth Barnes and Ariel Herbert-Voss and William Hebgen Guss and Alex Nichol and Alex Paino and Nikolas Tezak and Jie Tang and Igor Babuschkin and Suchir Balaji and Shantanu Jain and William Saunders and Christopher Hesse and Andrew N. Carr and Jan Leike and Josh Achiam and Vedant Misra and Evan Morikawa and Alec Radford and Matthew Knight and Miles Brundage and Mira Murati and Katie Mayer and Peter Welinder and Bob McGrew and Dario Amodei and Sam McCandlish and Ilya Sutskever and Wojciech Zaremba},
      year={2021},
      eprint={2107.03374},
      archivePrefix={arXiv},
      primaryClass={cs.LG},
      url={https://arxiv.org/abs/2107.03374}, 
}

@misc{austin2021mbpp,
      title={Program Synthesis with Large Language Models}, 
      author={Jacob Austin and Augustus Odena and Maxwell Nye and Maarten Bosma and Henryk Michalewski and David Dohan and Ellen Jiang and Carrie Cai and Michael Terry and Quoc Le and Charles Sutton},
      year={2021},
      eprint={2108.07732},
      archivePrefix={arXiv},
      primaryClass={cs.PL},
      url={https://arxiv.org/abs/2108.07732}, 
}

@misc{zheng2023mtbench,
      title={Judging LLM-as-a-Judge with MT-Bench and Chatbot Arena}, 
      author={Lianmin Zheng and Wei-Lin Chiang and Ying Sheng and Siyuan Zhuang and Zhanghao Wu and Yonghao Zhuang and Zi Lin and Zhuohan Li and Dacheng Li and Eric P. Xing and Hao Zhang and Joseph E. Gonzalez and Ion Stoica},
      year={2023},
      eprint={2306.05685},
      archivePrefix={arXiv},
      primaryClass={cs.CL},
      url={https://arxiv.org/abs/2306.05685}, 
}

@misc{nvidia2025nvidianemotronnano2,
      title={NVIDIA Nemotron Nano 2: An Accurate and Efficient Hybrid Mamba-Transformer Reasoning Model},
      author={NVIDIA},
      year={2025},
      eprint={2508.14444},
      archivePrefix={arXiv},
      primaryClass={cs.CL},
      url={https://arxiv.org/abs/2508.14444},
}

@article{huang2026step,
  title={Step 3.5 Flash: Open Frontier-Level Intelligence with 11B Active Parameters},
  author={Huang, Ailin and Li, Ang and Kong, Aobo and Wang, Bin and Jiao, Binxing and Dong, Bo and Wang, Bojun and Chen, Boyu and Li, Brian and Ma, Buyun and others},
  journal={arXiv preprint arXiv:2602.10604},
  year={2026}
}

@misc{ding2023ultrachat,
      title={Enhancing Chat Language Models by Scaling High-quality Instructional Conversations}, 
      author={Ning Ding and Yulin Chen and Bokai Xu and Yujia Qin and Zhi Zheng and Shengding Hu and Zhiyuan Liu and Maosong Sun and Bowen Zhou},
      year={2023},
      eprint={2305.14233},
      archivePrefix={arXiv},
      primaryClass={cs.CL},
      url={https://arxiv.org/abs/2305.14233}, 
}

@misc{li2024eagle2fasterinferencelanguage,
      title={EAGLE-2: Faster Inference of Language Models with Dynamic Draft Trees}, 
      author={Yuhui Li and Fangyun Wei and Chao Zhang and Hongyang Zhang},
      year={2024},
      eprint={2406.16858},
      archivePrefix={arXiv},
      primaryClass={cs.CL},
      url={https://arxiv.org/abs/2406.16858}, 
}

@article{hu2025bridging,
  title={Bridging Draft Policy Misalignment: Group Tree Optimization for Speculative Decoding},
  author={Hu, Shijing and Li, Jingyang and Lu, Zhihui and Zhou, Pan},
  journal={arXiv preprint arXiv:2509.22134},
  year={2025}
}

@article{miao2023specinfer,
  title={Specinfer: Accelerating generative large language model serving with tree-based speculative inference and verification},
  author={Miao, Xupeng and Oliaro, Gabriele and Zhang, Zhihao and Cheng, Xinhao and Wang, Zeyu and Zhang, Zhengxin and Wong, Rae Ying Yee and Zhu, Alan and Yang, Lijie and Shi, Xiaoxiang and others},
  journal={arXiv preprint arXiv:2305.09781},
  year={2023}
}

@misc{codealpaca,
  author = {Sahil Chaudhary},
  title = {Code Alpaca: An Instruction-following LLaMA model for code generation},
  year = {2023},
  publisher = {GitHub},
  journal = {GitHub repository},
  howpublished = {\url{https://github.com/sahil280114/codealpaca}},
}

@misc{zhang2025learningharmonizedrepresentationsspeculative,
      title={Learning Harmonized Representations for Speculative Sampling}, 
      author={Lefan Zhang and Xiaodan Wang and Yanhua Huang and Ruiwen Xu},
      year={2025},
      eprint={2408.15766},
      archivePrefix={arXiv},
      primaryClass={cs.LG},
      url={https://arxiv.org/abs/2408.15766}, 
}

@misc{du2024glide,
      title={GliDe with a CaPE: A Low-Hassle Method to Accelerate Speculative Decoding}, 
      author={Cunxiao Du and Jing Jiang and Xu Yuanchen and Jiawei Wu and Sicheng Yu and Yongqi Li and Shenggui Li and Kai Xu and Liqiang Nie and Zhaopeng Tu and Yang You},
      year={2024},
      eprint={2402.02082},
      archivePrefix={arXiv},
      primaryClass={cs.CL},
      url={https://arxiv.org/abs/2402.02082}, 
}

@misc{zheng2024sglangefficientexecutionstructured,
      title={SGLang: Efficient Execution of Structured Language Model Programs}, 
      author={Lianmin Zheng and Liangsheng Yin and Zhiqiang Xie and Chuyue Sun and Jeff Huang and Cody Hao Yu and Shiyi Cao and Christos Kozyrakis and Ion Stoica and Joseph E. Gonzalez and Clark Barrett and Ying Sheng},
      year={2024},
      eprint={2312.07104},
      archivePrefix={arXiv},
      primaryClass={cs.AI},
      url={https://arxiv.org/abs/2312.07104}, 
}

@misc{yan2025scalinglawsspeculativedecoding,
      title={Scaling Laws for Speculative Decoding}, 
      author={Siyuan Yan and Mo Zhu and Guo-qing Jiang and Jianfei Wang and Jiaxing Chen and Wentai Zhang and Xiang Liao and Xiao Cui and Chen Zhang and Zhuoran Song and Ran Zhu},
      year={2025},
      eprint={2505.07858},
      archivePrefix={arXiv},
      primaryClass={cs.CL},
      url={https://arxiv.org/abs/2505.07858}, 
}

@inproceedings{xia-etal-2024-unlocking,
    title = "Unlocking Efficiency in Large Language Model Inference: A Comprehensive Survey of Speculative Decoding",
    author = "Xia, Heming and Yang, Zhe and Dong, Qingxiu and Wang, Peiyi and Li, Yongqi  and Ge, Tao and Liu, Tianyu and Li, Wenjie and Sui, Zhifang",
    editor = "Ku, Lun-Wei and Martins, Andre and Srikumar, Vivek",
    booktitle = "Findings of the Association for Computational Linguistics ACL 2024",
    month = aug,
    year = "2024",
    address = "Bangkok, Thailand and virtual meeting",
    publisher = "Association for Computational Linguistics",
    url = "https://aclanthology.org/2024.findings-acl.456",
    doi = "10.18653/v1/2024.findings-acl.456",
    pages = "7655--7671",
}

@article{zhou2023distillspec,
  title={Distillspec: Improving speculative decoding via knowledge distillation},
  author={Zhou, Yongchao and Lyu, Kaifeng and Rawat, Ankit Singh and Menon, Aditya Krishna and Rostamizadeh, Afshin and Kumar, Sanjiv and Kagy, Jean-Fran{\c{c}}ois and Agarwal, Rishabh},
  journal={arXiv preprint arXiv:2310.08461},
  year={2023}
}

@article{liu2023online,
  title={Online speculative decoding},
  author={Liu, Xiaoxuan and Hu, Lanxiang and Bailis, Peter and Cheung, Alvin and Deng, Zhijie and Stoica, Ion and Zhang, Hao},
  journal={arXiv preprint arXiv:2310.07177},
  year={2023}
}

@misc{qwen3technicalreport,
      title={Qwen3 Technical Report}, 
      author={Qwen Team},
      year={2025},
      eprint={2505.09388},
      archivePrefix={arXiv},
      primaryClass={cs.CL},
      url={https://arxiv.org/abs/2505.09388}, 
}

@article{agarwal2025gpt,
  title={gpt-oss-120b \& gpt-oss-20b model card},
  author={Agarwal, Sandhini and Ahmad, Lama and Ai, Jason and Altman, Sam and Applebaum, Andy and Arbus, Edwin and Arora, Rahul K and Bai, Yu and Baker, Bowen and Bao, Haiming and others},
  journal={arXiv preprint arXiv:2508.10925},
  year={2025}
}

@article{angelslim2026,
  title={AngelSlim: A more accessible, comprehensive, and efficient toolkit for large model compression},
  author={Hunyuan AI Infra Team},
  journal={arXiv preprint arXiv:2602.21233},
  year={2026}
}

@misc{li2022diffusionlm,
      title={Diffusion-LM Improves Controllable Text Generation}, 
      author={Xiang Lisa Li and John Thickstun and Ishaan Gulrajani and Percy Liang and Tatsunori B. Hashimoto},
      year={2022},
      eprint={2205.14217},
      archivePrefix={arXiv},
      primaryClass={cs.CL},
      url={https://arxiv.org/abs/2205.14217}, 
}

@misc{strudel2022selfconditioned,
      title={Self-conditioned Embedding Diffusion for Text Generation}, 
      author={Robin Strudel and Corentin Tallec and Florent Altché and Yilun Du and Yaroslav Ganin and Arthur Mensch and Will Grathwohl and Nikolay Savinov and Sander Dieleman and Laurent Sifre and Rémi Leblond},
      year={2022},
      eprint={2211.04236},
      archivePrefix={arXiv},
      primaryClass={cs.CL},
      url={https://arxiv.org/abs/2211.04236}, 
}

@misc{austin2023structured,
      title={Structured Denoising Diffusion Models in Discrete State-Spaces}, 
      author={Jacob Austin and Daniel D. Johnson and Jonathan Ho and Daniel Tarlow and Rianne van den Berg},
      year={2023},
      eprint={2107.03006},
      archivePrefix={arXiv},
      primaryClass={cs.LG},
      url={https://arxiv.org/abs/2107.03006}, 
}

@misc{he2022diffusionbert,
      title={DiffusionBERT: Improving Generative Masked Language Models with Diffusion Models}, 
      author={Zhengfu He and Tianxiang Sun and Kuanning Wang and Xuanjing Huang and Xipeng Qiu},
      year={2022},
      eprint={2211.15029},
      archivePrefix={arXiv},
      primaryClass={cs.CL},
      url={https://arxiv.org/abs/2211.15029}, 
}

@misc{sahoo2024simple,
      title={Simple and Effective Masked Diffusion Language Models}, 
      author={Subham Sekhar Sahoo and Marianne Arriola and Yair Schiff and Aaron Gokaslan and Edgar Marroquin and Justin T Chiu and Alexander Rush and Volodymyr Kuleshov},
      year={2024},
      eprint={2406.07524},
      archivePrefix={arXiv},
      primaryClass={cs.CL},
      url={https://arxiv.org/abs/2406.07524}, 
}

@misc{shi2025simplified,
      title={Simplified and Generalized Masked Diffusion for Discrete Data}, 
      author={Jiaxin Shi and Kehang Han and Zhe Wang and Arnaud Doucet and Michalis K. Titsias},
      year={2025},
      eprint={2406.04329},
      archivePrefix={arXiv},
      primaryClass={cs.LG},
      url={https://arxiv.org/abs/2406.04329}, 
}

@article{li2025survey,
  title={A survey on diffusion language models},
  author={Li, Tianyi and Chen, Mingda and Guo, Bowei and Shen, Zhiqiang},
  journal={arXiv preprint arXiv:2508.10875},
  year={2025}
}

\appendix

\section{Appendix}

\subsection{Training Implementation}
\label{apd:train}

The draft models are optimized for 6 epochs using AdamW
with a learning rate of 6e$-$4
, a gradient clipping threshold of 1.0, and a cosine schedule with a warmup ratio of
0.04. We train on our training data mixture with a maximum
sequence length of 3072 tokens; for each sequence, 512 anchor positions are randomly sampled. For the progressive position-weighted loss introduced in Section~\ref{sec:progressive_loss}, we set the initial decay parameter to $\gamma_0 = 4.5$ and increase $\gamma$ by $1$ after each training epoch. For GPT-OSS-20B, we instead keep $\gamma$ fixed at $4$ throughout training, since its block size is only $8$ and the loss-weight schedule has a much smaller effect on later-position tokens in such a short block. Our main results use the full $2.4$\,M-sample training corpus with a global batch size of $64$. For ablation studies conducted on UltraChat and ShareGPT, as well as for the data-scaling comparison using the $800$K-sample Nemotron and CodeAlpaca mixture adopted by DFlash, we use a global batch size of $32$ while keeping all other optimization hyperparameters unchanged.  For every dataset, we keep only the original prompts and regenerate the responses with the corresponding target model using a sampling temperature of $0.6$; the draft model is then trained to predict only these regenerated response tokens, so that its training distribution matches the target model's own output distribution.

\subsection{Baselines}
\label{apd:baseline}
We compare \modelname with the vanilla autoregressive decoding (baseline) and autoregressive speculative
decoding method EAGLE3~\citep{li2025eagle} and state-of-the-art method DFlash~\citep{chen2026dflashblockdiffusionflash}. For EAGLE3 on Qwen3 models, we use the checkpoints released by AngelSlim~\citep{angelslim2026}. For DFlash, we use the official checkpoints released by DFlash team. For each task, we assess the performance of the draft models using average acceptance length $\tau$ and end-to-end decoding speedup over the autoregressive baseline. We conduct all experiments on NVIDIA H20 GPUs unless otherwise specified.

\subsection{Experiments compute resources}
\label{apd:compute_res}
\modelname draft models trained on the full 2.4\,M-sample corpus used 32 GPUs with distributed data parallelism. Under this setting, training the draft model for the Qwen3-8B target takes approximately $160$ GPU-hours per device (i.e., $\sim\!160$\,h wall-clock on 32 GPUs), the draft model for Qwen3-4B takes approximately $100$\,h, and the draft model for GPT-OSS-20B takes approximately $90$\,h. Inference-time evaluation is conducted on a single GPU for each target model, consistent with the deployment scenario of speculative decoding. A complete pass over the full evaluation benchmark takes approximately 4h on a single GPU for the Qwen3-8B target.

\subsection{Performance on SGLang}

\begin{table*}[h]
\centering
\begin{tabular}{ll ccccc}
\toprule
 & & \multicolumn{5}{c}{Concurrency} \\
\cmidrule(lr){3-7}
\multirow{-2}{*}{Task} & \multirow{-2}{*}{Method} & 1 & 4 & 8 & 16 & 32 \\
\midrule
\multirow{3}{*}{GSM8K} & Baseline & 162.5 & 610.8 & 1,058.8 & 1,046.9 & 1,053.7 \\
 & DFlash & 596.6 & 1,048.1 & 1,057.6 & 1,045.5 & 1,068.6 \\
 & \modelname & \textbf{642.6} & \textbf{1,764.8} & \textbf{1,815.2} & \textbf{1,828.7} & \textbf{1,809.3} \\
\midrule
\multirow{3}{*}{HumanEval} & Baseline & 163.6 & 620.9 & 1,158.2 & 1,273.7 & 1,212.0 \\
 & DFlash & \textbf{561.1} & 1,142.9 & 1,178.7 & 1,174.9 & 1,149.9 \\
 & \modelname & 554.5 & \textbf{1,538.6} & \textbf{1,650.2} & \textbf{1,672.1} & \textbf{1,796.9} \\
\bottomrule
\end{tabular}
\caption{Throughput (tokens/s) on SGLang with H20 GPU for Qwen3-8B under varying concurrency levels. Best results are in \textbf{bold}.}
\label{tab:sglang_h20}
\end{table*}

To evaluate the practical acceleration in realistic serving scenarios, we deploy \modelname on SGLang~\citep{zheng2024sglangefficientexecutionstructured} and measure end-to-end throughput under varying concurrency levels on an H20 GPU. As shown in Table~\ref{tab:sglang_h20}, \modelname consistently achieves the highest throughput across all concurrency levels on both GSM8K and HumanEval. As concurrency increases and the system becomes more compute-bound, the advantage of \modelname over DFlash grows more pronounced, demonstrating that our method scales favorably under higher load. 

\subsection{Detailed Results of the Ablation Study}
\label{apd:ablation}

In this section, we present the underlying data for the figures shown in Section~\ref{sec:draft_layers} and Sections~\ref{sec:analysis_features}.  We use ShareGPT and UltraChat200K~\citep{ding2023ultrachat} to do ablation study for comparison with Baseline Models. For the ablation study of number of draft layers, we use NVIDIA
Nemotron Post-Training Dataset V2~\citep{nvidia2025nvidianemotronnano2}, CodeAlpaca~\citep{codealpaca} which is the full DFlash training dataset. To prevent test set leakage, we removed all samples from the training set that had a 32-gram or higher overlap with the test set.

\begin{table*}[htbp]
\centering
\resizebox{\textwidth}{!}{%
\setlength{\tabcolsep}{2pt}
\begin{tabular}{llcccccccccccccc}
\toprule
\multirow{2}{*}{\normalsize{Method}} & \multirow{2}{*}{\normalsize{Layers}} & \multicolumn{6}{c}{MATH} & \multicolumn{4}{c}{CODE} & \multicolumn{2}{c}{CHAT} & \multicolumn{2}{c}{\multirow{2}{*}{Avg.}} \\
\cmidrule(lr){3-14}
& & \multicolumn{2}{c}{\textit{GSM8K}} & \multicolumn{2}{c}{\textit{MATH500}} & \multicolumn{2}{c}{\textit{AIME25}} & \multicolumn{2}{c}{\textit{HumanEval}} & \multicolumn{2}{c}{\textit{MBPP}} & \multicolumn{2}{c}{\textit{MT-Bench}} & \multicolumn{2}{c}{} \\
\midrule
\multicolumn{2}{c}{Temperature = 0} & \textit{Speedup} & \textit{$\tau$} & \textit{Speedup} & \textit{$\tau$} & \textit{Speedup} & \textit{$\tau$} & \textit{Speedup} & \textit{$\tau$} & \textit{Speedup} & \textit{$\tau$} & \textit{Speedup} & \textit{$\tau$} & \textit{Speedup} & \textit{$\tau$} \\
\midrule

\multirow{3}{*}{DFlash} 
& 5 Layers          & 3.63 & 4.47 & 3.50 & 4.28 & 3.39 & 4.14 & 3.45 & 4.25 & 3.11 & 3.84 & 2.78 & 3.78 & 3.31 & 4.13 \\
& 7 Layers          & 3.70 & 4.55 & 3.54 & 4.34 & 3.45 & 4.19 & 3.51 & 4.33 & 3.17 & 3.88 & \textbf{2.81} & 3.80 & 3.36 & 4.18 \\
& 9 Layers          & 3.73 & 4.59 & 3.53 & 4.33 & \textbf{3.52} & 4.30 & 3.53 & 4.34 & 3.11 & 3.81 & 2.78 & 3.78 & 3.37 & 4.19 \\
\midrule
\multirow{3}{*}{\modelname} 
& 5 Layers          & 3.72 & 4.83 & 3.49 & 4.52 & 3.38 & 4.36 & 3.50 & 4.55 & 3.14 & 4.07 & 2.73 & 3.92 & 3.33 & 4.38 \\
& 7 Layers          & 3.70 & 4.82 & 3.51 & 4.53 & 3.49 & 4.48 & 3.53 & 4.59 & \textbf{3.19} & \textbf{4.16} & 2.73 & 3.92 & 3.36 & 4.42 \\
& 9 Layers          & \textbf{3.80} & \textbf{4.94} & \textbf{3.57} & \textbf{4.67} & 3.51 & \textbf{4.52} & \textbf{3.55} & \textbf{4.62} & 3.17 & \textbf{4.16} & 2.76 & \textbf{4.00} & \textbf{3.39} & \textbf{4.49} \\

\bottomrule
\end{tabular}%
}
\caption{Detailed ablation study results on Qwen3-4B about number of target layers across mathematical reasoning, code generation, and conversation benchmarks. Speedup denotes wall-clock speedup ratio and $\tau$ denotes acceptance length. Best results in each category are in \textbf{bold}.}
\label{tab:target_results}
\end{table*}

\begin{table*}[htbp]
\centering
\resizebox{\textwidth}{!}{%
\setlength{\tabcolsep}{2pt}
\begin{tabular}{llcccccccccccccc}
\toprule
\multirow{2}{*}{\normalsize{Method}} & \multirow{2}{*}{\normalsize{Layers}} & \multicolumn{6}{c}{MATH} & \multicolumn{4}{c}{CODE} & \multicolumn{2}{c}{CHAT} & \multicolumn{2}{c}{\multirow{2}{*}{Avg.}} \\
\cmidrule(lr){3-14}
& & \multicolumn{2}{c}{\textit{GSM8K}} & \multicolumn{2}{c}{\textit{MATH500}} & \multicolumn{2}{c}{\textit{AIME25}} & \multicolumn{2}{c}{\textit{HumanEval}} & \multicolumn{2}{c}{\textit{MBPP}} & \multicolumn{2}{c}{\textit{MT-Bench}} & \multicolumn{2}{c}{} \\
\midrule
\multicolumn{2}{c}{Temperature = 0} & \textit{Speedup} & \textit{$\tau$} & \textit{Speedup} & \textit{$\tau$} & \textit{Speedup} & \textit{$\tau$} & \textit{Speedup} & \textit{$\tau$} & \textit{Speedup} & \textit{$\tau$} & \textit{Speedup} & \textit{$\tau$} & \textit{Speedup} & \textit{$\tau$} \\
\midrule

\multirow{3}{*}{DFlash} 
& 5 Layers & 5.30 & 6.57 & 6.17 & 7.70 & 5.97 & 7.40 & 4.61 & 5.71 & 4.66 & 5.85 & 3.14 & 4.58 & 4.98 & 6.30 \\
& 6 Layers & 5.38 & 6.82 & 6.27 & 8.03 & 6.30 & 7.89 & 4.66 & 5.92 & 4.70 & 6.04 & 3.11 & 4.71 & 5.07 & 6.57 \\
& 7 Layers & 5.43 & 7.06 & 6.31 & 8.26 & 5.98 & 7.78 & 4.60 & 5.98 & 4.65 & 6.12 & 3.11 & 4.80 & 5.01 & 6.67 \\
\midrule
\multirow{3}{*}{\modelname} 
& 5  Layers & 5.42 & 6.85 & 6.35 & 8.03 & 6.16 & 7.68 & \textbf{4.73} & 5.88 & \textbf{4.81} & 6.10 & \textbf{3.22} & 4.77 & 5.12 & 6.55 \\
& 6 Layers & 5.56 & 7.13 & 6.32 & 8.19 & 6.23 & 7.96 & 4.71 & 6.03 & 4.77 & 6.21 & 3.16 & 4.79 & 5.13 & 6.72 \\
& 7 Layers & \textbf{5.58} & \textbf{7.29} & \textbf{6.37} & \textbf{8.41} & \textbf{6.36} & \textbf{8.27} & 4.70 & \textbf{6.16} & 4.80 & \textbf{6.38} & 3.15 & \textbf{4.88} & \textbf{5.16} & \textbf{6.90} \\

\bottomrule
\end{tabular}%
}
\caption{Detailed ablation study results on Qwen3-4B across different draft layer settings across mathematical reasoning, code generation, and conversation benchmarks. Speedup denotes wall-clock speedup ratio and $\tau$ denotes acceptance length. Best results in each category are in \textbf{bold}.}
\label{tab:draft_layer_ablation}
\end{table*}

\subsection{Scaling Training Data}
Detailed results on \textbf{Qwen3-8B} across different training data volumes in Figure~\ref{fig:speedup}. 
\begin{table*}[htbp]
\centering
\resizebox{\textwidth}{!}{%
\setlength{\tabcolsep}{2pt}
\begin{tabular}{llcccccccccccccc}
\toprule
\multirow{2}{*}{\normalsize{Method}} & \multirow{2}{*}{\normalsize{Data}} & \multicolumn{6}{c}{MATH} & \multicolumn{4}{c}{CODE} & \multicolumn{2}{c}{CHAT} & \multicolumn{2}{c}{\multirow{2}{*}{Avg.}} \\
\cmidrule(lr){3-14}
& & \multicolumn{2}{c}{\textit{GSM8K}} & \multicolumn{2}{c}{\textit{MATH500}} & \multicolumn{2}{c}{\textit{AIME25}} & \multicolumn{2}{c}{\textit{HumanEval}} & \multicolumn{2}{c}{\textit{MBPP}} & \multicolumn{2}{c}{\textit{MT-Bench}} & \multicolumn{2}{c}{} \\
\midrule
\multicolumn{2}{c}{Temperature = 0} & \textit{Speedup} & \textit{$\tau$} & \textit{Speedup} & \textit{$\tau$} & \textit{Speedup} & \textit{$\tau$} & \textit{Speedup} & \textit{$\tau$} & \textit{Speedup} & \textit{$\tau$} & \textit{Speedup} & \textit{$\tau$} & \textit{Speedup} & \textit{$\tau$} \\
\midrule

\multirow{3}{*}{\modelname} 
& 270K & 4.40 & 5.70 & 4.00 & 5.19 & 3.75 & 4.81 & 3.87 & 4.98 & 3.58 & 4.63 & 2.84 & 4.09 & 3.74 & 4.90 \\
& 800K & 5.35 & 6.94 & 6.23 & 8.23 & 5.97 & 7.74 & 4.65 & 6.01 & 4.59 & 6.03 & 3.02 & 4.61 & 4.97 & 6.59 \\
& 2.4M & \textbf{6.02} & \textbf{7.88} & \textbf{6.72} & \textbf{8.95} & \textbf{6.21} & \textbf{8.12} & \textbf{5.42} & \textbf{7.08} & \textbf{5.16} & \textbf{6.86} & \textbf{3.23} & \textbf{5.09} & \textbf{5.46} & \textbf{7.33} \\

\bottomrule
\end{tabular}%
}
\caption{Detailed results on \textbf{Qwen3-8B} across different training data volumes. Speedup denotes wall-clock speedup ratio and $\tau$ denotes acceptance length. Best results in each category are in \textbf{bold}.}
\label{tab:data_scaling_ablation}
\end{table*}

\subsection{Software Dependencies}
Our codebase is implemented in Python, primarily relying on \texttt{torch} (v2.9.1), \texttt{transformers} (v4.57.1), \texttt{datasets} (v4.8.4), \texttt{sglang} (v0.5.6), and \texttt{numpy} (v2.4.3).
\end{document}